\documentclass[a4paper,fleqn]{cas-dc}

\usepackage[numbers]{natbib}
\usepackage{makecell}
\newcommand{\etal}{\textit{et al.}\xspace}

\begin{document}
\shortauthors{Z. Zhang et~al.}
\shorttitle{SPAST}
\title [mode = title]{SPAST: Arbitrary Style Transfer with Style Priors via Pre-trained Large-scale Model}

\author[1]{Zhanjie Zhang}[orcid=0000-0002-8966-1328]
\author[1]{Quanwei Zhang}
\author[1]{Junsheng Luan}   
\author[1]{Mengyuan Yang}
\author[2]{Yun Wang}
\author[1]{Lei Zhao\corref{cor1}}
                
\affiliation[1]{College of Computer Science and Technology, Zhejiang University, No. 38, Zheda Road, Hangzhou 310000, China}
\affiliation[2]{Department of Computer Science, City University of Hong Kong, Tat Chee Avenue, Kowloon, Hong Kong SAR}

\cortext[cor1]{Corresponding author}
\nonumnote{E-mail addresses: cszzj@zju.edu.cn (Z. Zhang), cszqw@zju.edu.cn (Q. Zhang), l.junsheng121@zju.edu.cn (J. Luan), yangmy412@zju.edu.cn (M. Yang), ywang3875-c@my.cityu.edu.hk (Y.Wang), cszhl@zju.edu.cn (L. Zhao)}

\begin{abstract}
	Given an arbitrary content and style image, arbitrary style transfer aims to render a new stylized image which preserves the
content image’s structure and possesses the style image’s style. Existing arbitrary style transfer methods are based on either
small models or pre-trained large-scale models. The small model-based methods fail to generate high-quality stylized images,
bringing artifacts and disharmonious patterns. The pre-trained large-scale model-based methods can generate high-quality
stylized images but struggle to preserve the content structure and cost long inference time. To this end, we propose a new
framework, called SPAST, to generate high-quality stylized images with less inference time. Specifically, we design a novel
Local-global Window Size Stylization Module (LGWSSM) to fuse style features into content features. Besides, we introduce
a novel style prior loss, which can dig out the style priors from a pre-trained large-scale model into the SPAST and motivate
the SPAST to generate high-quality stylized images with short inference time.We conduct abundant experiments to verify that
our proposed method can generate high-quality stylized images and less inference time compared with the SOTA arbitrary
style transfer methods.
\end{abstract}


\maketitle

\section{Introduction}
	Arbitrary style transfer aims to render a stylized image
which preserves the content image’s structure and possesses
the style of the style image. Existing arbitrary style transfer
methods can be generally divided into small model-based
methods (SMM) and pre-trained large-scale model-based
methods (LMM).
	\begin{figure}[t]
		\centering
		\includegraphics[width=1\columnwidth]{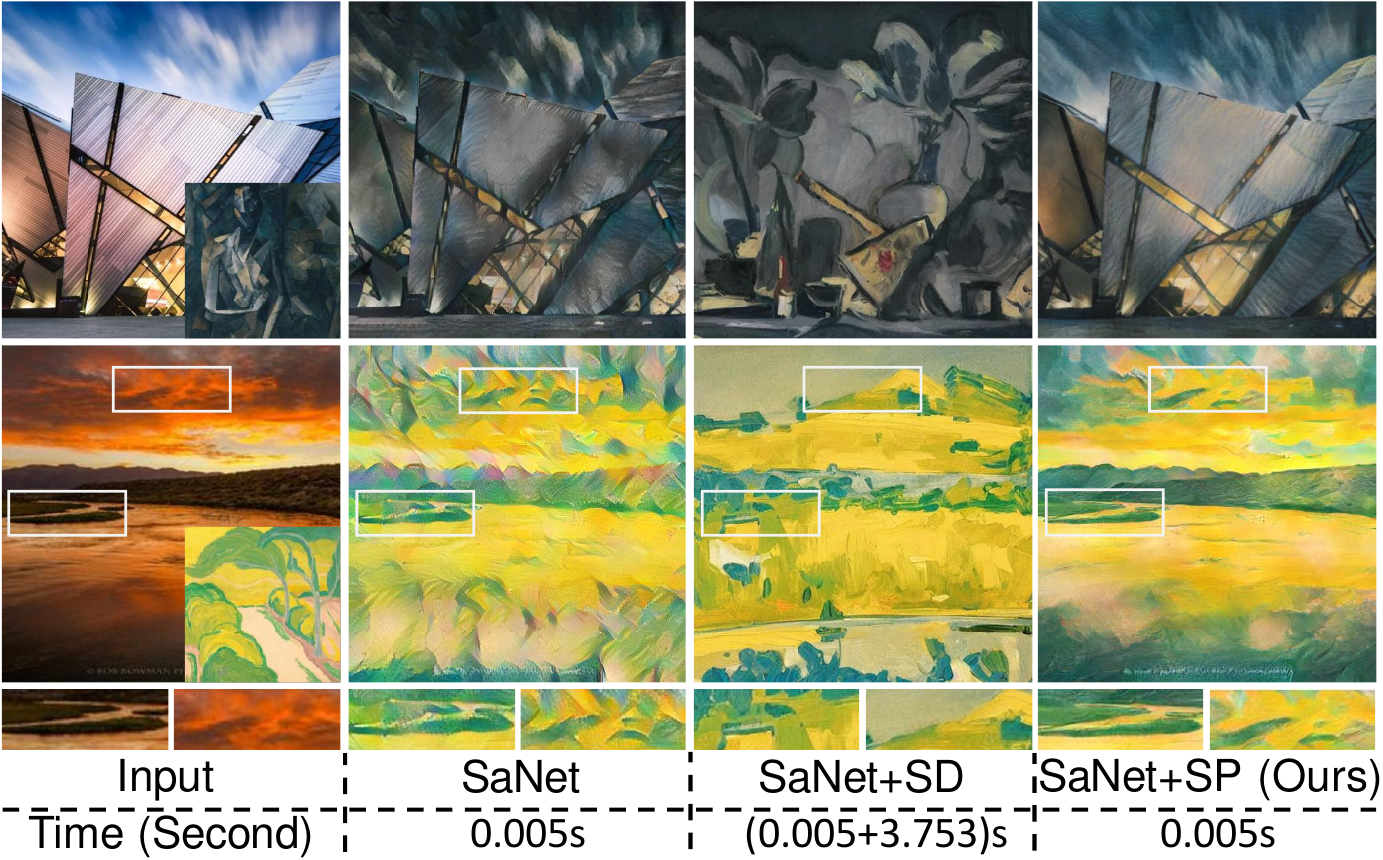} 
		\caption{ The first column shows the input images. The
other three columns show the stylized images produced by
SaNet~\protect\cite{park2019arbitrary}, SaNet+SD~\protect\cite{rombach2022high}, and SaNet+SP (Ours).}
		\label{visual}
	\end{figure}
	
	Specifically, the SMM methods~\cite{liu2021adaattn,deng2020arbitrary,chen2021artistic,park2019arbitrary,deng2022stytr2,wang2022aesust,wang2022fine,gatys2015neural,an2021artflow,hong2023aespa,zhu2023all,yang2022gating,kim2021uncorrelated,ma2020image,xue2021cycle,zhang2023csast} focus on design a small-scale network
and learn style knowledge from less data. Firstly,
	Huang~\etal~\cite{huang2017arbitrary} trained a small-scale network to learn the mean and variance of deep image features to represent style
information. 
	Park~\etal~\cite{park2019arbitrary} first proposed to learn local style
feature patches via cross attention. Liu ~\etal~\cite{liu2021adaattn} proposed
attention-based normalization to further support style transfer. Deng ~\etal~\cite{deng2022stytr2} proposed a transformer-based network to fuse global style information into content features. Zhu ~\etal~\cite{zhu2023all} proposed all-to-key attention to learn relationship
between each position of content feature and each positions
of style features. However these methods always introduces
disharmonious patterns and evident artifacts (e.g., $2^{nd}$ col of Fig.~\ref{visual}).
	
	The pre-trained LMM~\cite{cheng2023general,nichol2021improved,zhang2022inversion,dhariwal2021diffusion,rombach2022high,yang2023zero,kwon2023diffusion,zhang2024artbank} can
generate high-quality stylized images since they learn massive
style knowledge from large amounts of data. Recently, 
	Parmar ~\etal~\cite{parmar2023zero} discovered editing directions from the
text embedding space to generate desired stylized images. 
	Kwon ~\etal~\cite{kwon2023diffusion} proposed to guide the generation process of  DDPM~\cite{ho2020denoising} via a pre-trained VIT~\cite{vaswani2017attention}. Zhang ~\etal~\cite{zhang2022inversion} proposed to learn style knowledge from example and guided SD~\cite{rombach2022high} to support style transfer. Cheng ~\etal~\cite{cheng2023general} proposed
a content-style inversion to extract contents and style. Although
these pre-trained LLM can generate high-quality
stylized images, they struggle to preserve the content
image’s structure well and cost long inference time (e.g., In $3^{rd}$ col of Fig.~\ref{visual}, simply feeding the stylized image from SaNet into pre-trained large-scale model Stable Diffusion (SD)~\cite{rombach2022high} fails to preserve content structure well and the inference time in such way will be the total time of SaNet and SD.). 
	
	To address these problems, we propose a novel method,
SPAST, which can generate high-quality stylized images,
preserve the content image’s structure well and cost less
inference time. Firstly, we argue that existing style transfer
methods~\cite{park2019arbitrary,zhu2023all,chen2016fast} based on small-scale models implement style transfer through the similarity of local contents between the content images and the style images, decreasing the quality of the stylized images.
	To this end, we propose a Local-global Window Size Stylization Model (LGWSSM) to fuse the style features into content features. LGWSSM pays attention on all positions and regions of style images. Besides, although simply feeding the stylized image generated by a small model-based method into a pre-trained large-scale model (Stable Diffusion~\cite{rombach2022high}) can generate high quality
stylized images, it still fails to preserve the content
structure well (e.g., $3^{rd}$ col of Fig.~\ref{visual}) and cost long inference time. Besides, Stable Diffusion can only generate the desired stylized images based on the text description instead of the style image. To this end, we fine-tune a pre-trained large-scale Artistic Stable Diffusion, which is trained on Wikiart~\cite{wikiart} and supports the use of a style image as a reference instead of a text description. Further, we propose a style prior (SP) loss that can dig out the abundant prior knowledge from our ASD  into SPAST. Our proposed style prior loss can not only improve the quality of stylized images generated by SPAST but also be applied to existing small model-based style transfer methods to improve the quality of stylized images (e.g., see in $4^{th}$ col of Fig.~\ref{visual}).
	The main contributions of this work are fourfold:
	\begin{itemize}
		\item We propose a novel framework, called \emph{SPAST}, which can dig out the prior knowledge from pre-trained large-scale models into small-scale model, generating high-quality stylized images with better content structure preservation and less inference time.
		\item A novel \emph{Local-global Window Size Stylization Module (LGWSSM)} is proposed to fuse the style features into content features and pay attention on all positions and regions of style images, which helps improve the quality of stylized images effectively.
		\item We fine-tune a pre-trained large-scale \emph{Artistic Stable Diffusion (ASD)} and introduce a novel optimization objective, called \emph{style prior} loss, which helps improve the quality of stylized images generated by \emph{SPAST} without increasing the inference time. 
		\item The abundant qualitative and quantitative experiments
verify that our proposed method can generate higher
quality stylized images than the SOTA methods.
	\end{itemize}
	
	\section{Related Work}
	\textbf{Small model-based methods (SSM).} 
	The SMM enable
style transfer with a small amount of data and parameters. For example, Gatys ~\etal\cite{gatys2015neural} pioneered the neural style transfer by iteratively minimizing the joint content and style loss in the feature space of a pre-trained deep neural network (VGG)~\cite{simonyan2014very}.  Chen ~\etal~\cite{chen2016fast} swapped each content feature with its closest-matching style feature patch.  Recently, some methods utilized adversarial loss~\cite{wang2022aesust,chen2021artistic,zhang2022domain} between style images and stylized images, improving the quality of stylized images. Yi~\etal~\cite{yi2024aesstyler} proposed to introduce style characteristics through aesthetic evaluation in the training process. Wang~\etal~\cite{wang2022aesust} proposed a two-stage strategy to inject the style information learned by the discriminator into the generator. Wang~\etal~\cite{wang2023stylediffusion} proposed a content-style disentanglement loss in the clip space, which can learn to transfer the style information of the style image onto the content image.
Although they can efficiently learn style information from
the style image, the small model-based methods usually
bring disharmonious patterns and obvious artifacts since
they are trained on a limited amount of data and parameters.
	
	\textbf{Pre-trained large-scale model-based methods (LLM)}. 
	The pre-trained LMM possess abundant prior knowledge
and can generate high-quality images. For example, Yue ~\etal\cite{yue2023dif} proposed to fuse style
feature into content feature via from multi-channel perspective
to generate the stylized image directly. He ~\etal\cite{he2023cartoondiff} proposed a zero-shot approach that generates stylized images
via diffusion transformer models. Zhang~\etal\cite{zhang2025dyartbank} proposed a dynamic implicit style prompt bank to learn style patterns from multiple artworks; this dynamic implicit style prompt bank can condition pre-trained large-scale generate stylized images with different styles. Liu~\etal\cite{liu2024intrinsic} proposed to balance intrinsic-external style distribution to create stylized images.
Cao~\etal~\cite{cao2025relactrl} introduced style information via degree of preference for style information between different modules.
Poole~\etal~\cite{poole2022dreamfusion} proposed
a Score Distillation Sampling method by minimizing the
KL divergence between Gaussian distribution with shred
means based on the forward process of diffusion and the
score functions. Gao~\etal~\cite{gao2024styleshot} introduced multiple expert heads to learn style information from different image resolutions and utilize this multi-scale style information to condition diffusion model to generate stylized images.
Although pre-trained LLM can generate
high-quality stylized images, they struggle to preserve the
content image’s structure well and cost long inference time.
	
	\section{Proposed Method}
\begin{figure*}[htb]
		\centering
		\includegraphics[width=2.09\columnwidth]{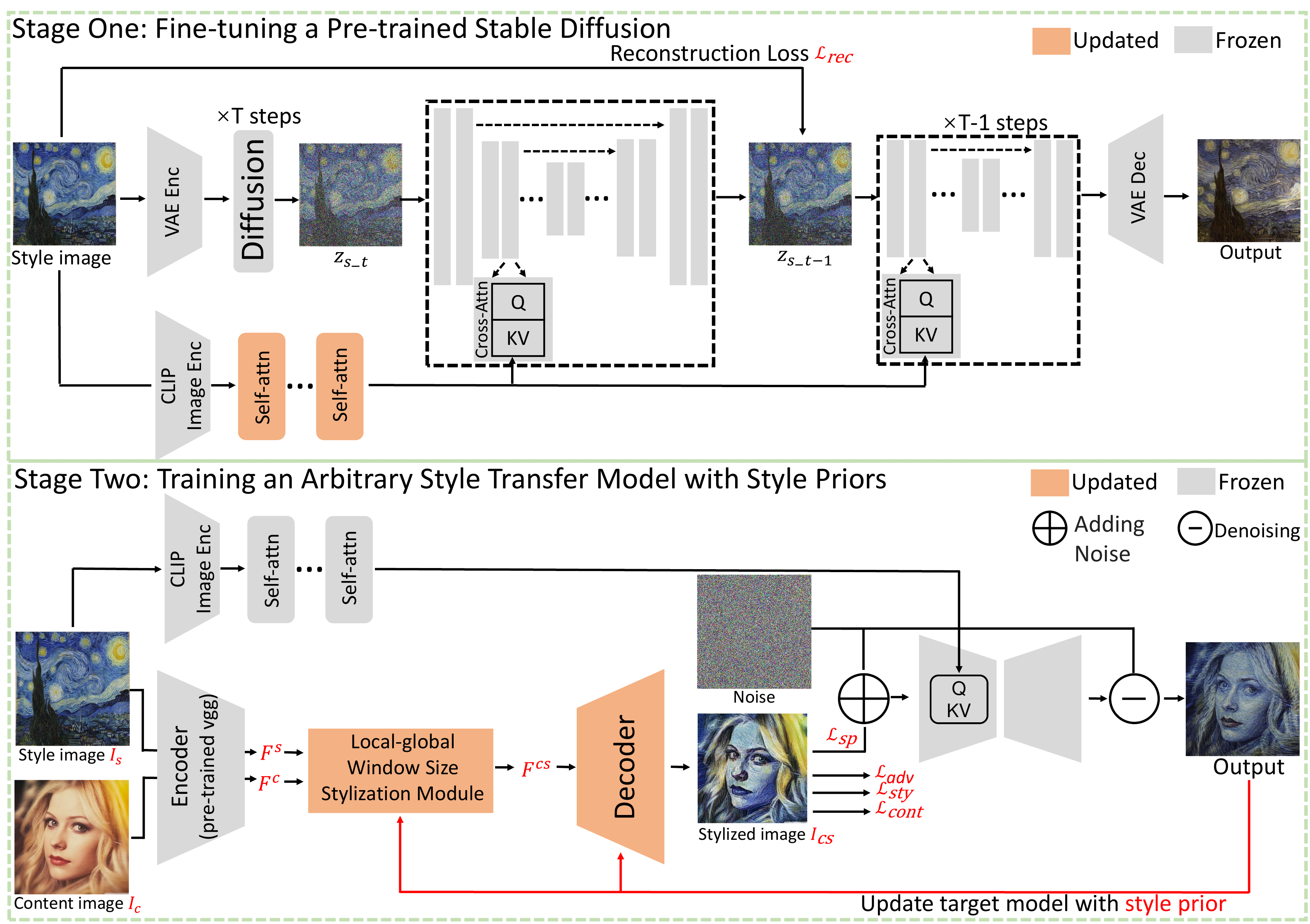} 
		\caption{Overview of our SPAST, which consists of two stages. Stage one: finetuning a pre-trained Stable Diffusion, obtaining an Artistic Stable Diffusion. Stage two: Training an arbitrary style transfer model with style priors.  
		}
		\label{image4}
	\end{figure*}

	The overview of our proposed SPAST is illustrated in Fig.~\ref{image4}. As we can see, the pipeline of our proposed SPAST consists of two stages: Fine-tuning a pre-trained Stable Diffusion to obtain an Artistic Stable Diffusion (ASD) and training an arbitrary style transfer method with style priors from ASD.
	
	In stage one, we fine-tune the Stable Diffusion~\cite{rombach2022high} on the given style image $I_{s}$ from Wikiart~\cite{wikiart}, obtaining an Artistic Stable Diffusion which supports to the use of style image as reference. 
	
	In stage two, we utilize the encoder $V_{vgg}$ to extract the feature from the content image (hence referred to as content features $F_{c}$) and the feature from style image (thus referred to as style features $F_{s}$). Further, the Local-global Window Size Stylization Module ($LGWSSM$) is used to fuse style feature into content feature, obtaining stylized feature $F_{cs}^{lg}$. Finally, we pull $F_{cs}^{lg}$ into the decoder $D$ to generate the stylized image $I_{cs}$. It is worth noting that the structure of the decoder here follows the SaNet~\cite{park2019arbitrary} and adopts the mirror structure of VGG.
	
	\begin{figure}[htb]
		\centering
		\includegraphics[width=1\columnwidth]{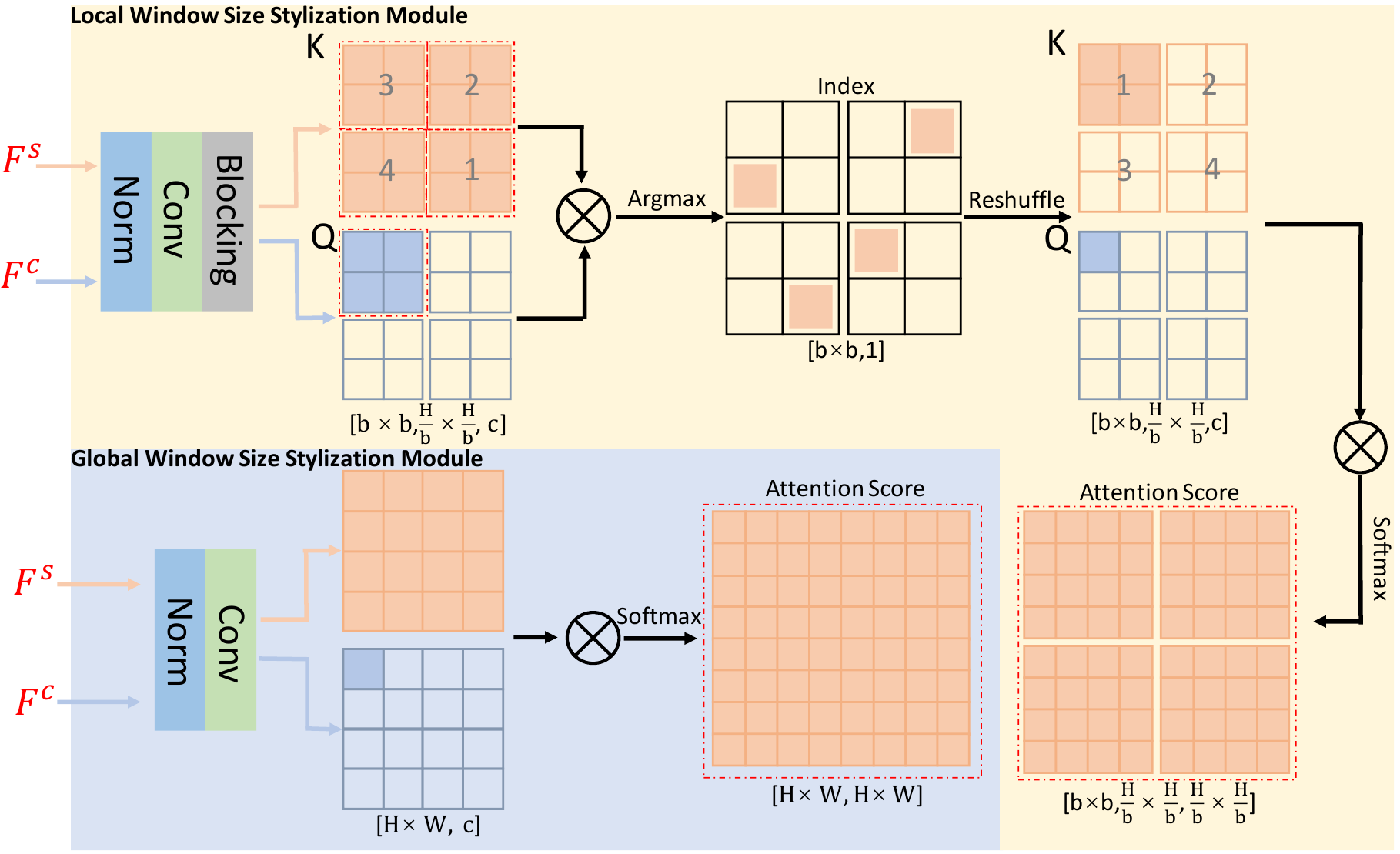} 
		\caption{Illustration of Local-global Window Size Stylization Module (LGWSSM).}
		\label{image7}
	\end{figure}
	\subsection{Local-global Window Size Stylization Module}
	\label{lgwssm}
	As shown in Fig.~\ref{image7}, our proposed Local-global Window Size Stylization Module ($LGWSSM$) consists of a Local Window Size Stylization Module ($LWSSM$) and a Global Window Size Stylization Module ($GWSSM$).
	

	\textbf{Local Window Size Stylization Module.}
	Firstly, the $LGWSSM$ finds the style image region that is closest to the content image region which is implemented as region-wise attention.
	It takes a region as a token instead of a specific position. The content feature $F_{c}$ and the style feature $F_{s}$ (i.e., $F_{c}$ and $F_{s}$$\in$ $R^{H, W, C}$) are spatially blocked into the non-overlapping region tensor: $Q_{b}$ and $K_{b}$.
	\begin{equation}
	\begin{aligned}
	&Q_{b}=B(N(F_c)), 
	K_{b}=B(N(F_s)),
	V_{b}=B(F_s)&\\
	\end{aligned}
	\end{equation}
	where $N(*)$ denotes the mean-variance channel-wise normalization, as used in instance normalization. The $B(*)$ denotes $1\times 1$ Conv-Blocking operation. The $Q_{b}$, $K_{b}$ and $V_{b}$ denotes $b\times b$ non-overlapping regions (i.e., $Q_{b}$, $K_{b}$ and $V_{b}$ $\in$ $R^{b\times b, \frac{H}{b} \times \frac{W}{b}, C}$).
	$Argmax$ is used to match only the most similar coarse-grained region, and the output of this step is the index matrix stores sparse indices of region-wise token across $b$$\times$$b$ regions:
	\begin{equation}
	L_{i d x}=\operatorname{argmax}\left(L\left(Q_{b}, K_{b}\right)\right),
	\end{equation}
	where $L$ means to search for the similar coarse-grained region. With $L_{i d x}$ (i.e., $L_{i d x}$$\in$ $R^{b\times b}$ ), we can rearrange the tokens of $K_{b}$ and $V_{b}$ to semantically matching the spatial arrangement of the tokens of $Q_{b}$:
	\begin{equation}
	\begin{aligned}
	\widetilde{K}_{b} & =\operatorname{rearrange}\left(K_{b}, L_{i d x}\right), \\
	\widetilde{V}_{b} & =\operatorname{rearrange}\left(V_{b}, L_{i d x}\right),
	\end{aligned}
	\end{equation}
	where $rearrange(*,*)$ denotes the rearrange operation. 
	We use $Q_{b}^{n}$, $K_{b}^{n}$ and $V_{b}^{n}$ to denote the $n$-th block of $Q_{b}$, $K_{b}$ and $V_{b}$ respectively (i.e., $Q_{b}^{n}$, $K_{b}^{n}$ and $V_{b}^{n}$ $\in$ $R^{\frac{H}{b} \times \frac{W}{b}, C}$ ).
	The $n$-th region-wise attention score $A_{b}^{n}$ is calculated from $Q_{b}^{n}$ and $\widetilde{K}_{b}^{n}$:
	\begin{equation}
	A_{b}^{n}=Softmax\left(Q_{b}^{n} \otimes \tilde{K}_{b}^{n}\right),
	\end{equation}
	where $\otimes$ denotes the matrix multiplication. The region-wise attention score matrix $A_{b}^{n}$, with the size of ($ \frac{H}{b} \times \frac{W}{b}, \frac{H}{b} \times \frac{W}{b}$), stores the sparse similarity correspondence of point-wise tokens within regions with the size of $\frac{H}{b} \times \frac{W}{b}$ indexed in the range $b\times b$. 
	
	Secondly, we compute region-based attention-weighted mean $M_{b}^{n}$ and region-based attention-weighted variance $S_{b}^{n}$ as below:
	\begin{equation}
	M_{b}^{n}=\widetilde{V}_{b}^{n} \otimes {A_{b}^{n}}^{\top},
	\end{equation}
	\begin{equation}
	S_{b}^{n}=\sqrt{(\widetilde{V}_{b}^{n} \cdot \widetilde{V}_{b}^{n}) \otimes {A_{b}^{n}}^{\top}- M_{b}^{n} \cdot M_{b}^{n}},
	\end{equation}
	where $\cdot$ denotes element-wise product. Then, each stylized feature region $F_{cs}^{n}$ can be computed as below:
	\begin{equation}
	F_{cs}^{n}=S_{b}^{n} \cdot N (F_c^{n}) + M_{b}^{n},
	\end{equation}
	After computing each $F_{cs}^{n}$, we can get the region-based stylized feature $F_{cs}^{b}$
	($F_{cs}^{b}$ consists of each stylized feature region: $F_{cs}^{1}$, $F_{cs}^{2}$ $\cdots$ $F_{cs}^{n}$).
	
	\textbf{Global Window Size Stylization Module.} To compute global window size attention map $A$, we formulate $Q(query)$, $K(key)$, $V(value)$ as:
	\begin{equation}
	\begin{aligned}
	Q & =f\left({N}\left(F_c^{}\right)\right), 
	K & =g\left({N}\left(F_s^{}\right)\right), 
	V & =h\left(F_s\right),
	\end{aligned}
	\end{equation}
	where $f$, $g$ and $h$ are 1$\times$1 learnable convolution layers, $Norm$ represents channel-wise mean-variance normalization. The attention map $A$ can be calculated as:
	\begin{equation}
	A={Softmax}\left(Q^{\top} \otimes K\right),
	\end{equation}
	where $\otimes$ denotes matrix multiplication. The attention-weighted mean can be computed as below:
	\begin{equation}
	M=V \otimes A^{\top}
	\end{equation}
	where $A \in R^{H W \times H W}$ and $V \in R^{H\times W , C}$. Then attention-weighted standard deviation $S \in R^{H\times W , C}$ as:
	\begin{equation}
	S=\sqrt{(V \cdot V) \otimes A^{\top}-M \cdot M},
	\end{equation}
	where $\cdot$ denotes element-wise product, corresponding scale in $S$ and shift in $M$ are used to generated transformed feature map:
	\begin{equation}
	F_{c s}=S \cdot N\left(F_c\right)+M,
	\end{equation}
	\textbf{Feature Transformation.} With the local window size stylization feature $F_{cs}^{b}$ and global window size stylization feature $F_{cs}$, the local-global window size stylization feature can be computed as below:
	\begin{equation}
	F_{cs}^{lg}=Unblock(F_{cs}^{b}) + F_{cs},
	\end{equation}
	where $Unblock(*)$ denotes the Unblocking-Conv operation. Finally, we input $F_{cs}^{lg}$ into the decoder $D$ to generate the stylized image $I_{cs}$:
	\begin{equation}
	I_{cs}=D(F_{cs}^{lg}).
	\end{equation}
	
	\subsection{Style Prior Loss}
	\label{3.1}
	Simply feeding the stylized image generated by a small model into the pre-trained large-scale model is an intuitive way to dig out the prior knowledge from a pre-trained large-scale model (e.g., Stable Diffusion~\cite{rombach2022high}). The stylized image by such a naive method shows the content structure degradation (e.g., $3^{rd}$ col of Fig.~\ref{visual}). To this end, we propose an Artistic Stable Diffusion based on Wikiart~\cite{wikiart}, as shown in the top half of Fig.~\ref{image4}. Specifically, given a style image $I_{s}$ from Wikiart, we first use VAE encoder to obtain latent style feature ${z_{s}}$ and then add noise ${\epsilon} \sim \mathcal{N}(0, {I})$ to obtain the noisy style feature ${z_{s}}_{\_ t}$ as below:
	\begin{equation}
	\label{noisy}
	{z_{s}}_{\_ t}=\sqrt{\bar{\alpha}_t} {I_{s}}+\sqrt{1-\bar{\alpha}_t} {\epsilon},
	\end{equation}
	Then, ${z_{s}}_{\_ t}$ is feed into Stable Diffusion~\cite{rombach2022high}. Besides, we utilize the CLIP~\cite{radford2021learning} image encoder to extract style information and utilize learnable self-attention network to further obtain style embedding $Attn(I_{s})$ which can condition pre-trained Stable Diffusion to reconstruct the style image. We train self-attention using the following reconstruction loss:
	\begin{equation}
	\mathcal{L}_{\text {rec }}=\mathbb{E}_{z, x, t}\left[\left\|\epsilon-\epsilon_\theta\left(z_{s\_t},Attn(I_{s}), t\right)\right\|_2^2\right],
	\end{equation}
	After above process, we get Artistic Stable Diffusion which possesses massive style priors and support the use of style images as reference. Further, we propose a style prior loss, which can dig out the massive style priors from Artistic Stable Diffusion into SPAST. Specifically, given a stylized image $I_{cs}$ generated by SPAST, we first use VAE encoder to obtain the latent stylized feature ${z_{cs}}$ and then apply noise to it with Eq.~\ref{noisy}, obtaining noisy stylized feature ${z_{cs}}_{\_t}$. Then, we feed ${z_{cs}}_{\_ t}$ into fixed Artistic Stable Diffusion and condition it to use corresponding style image $I_{s}$. We define the above process as an optimization objective as below:
	\begin{equation}
	\label{sp}
	\begin{aligned}
	\mathcal{L}_{\text {sp }}=\mathbb{E}_{t, \epsilon}[w(t) {\left(\hat{\epsilon}_\phi\left({z}_{cs\_t} ; Attn(I_{s}), t\right)-\epsilon\right)}],
	\end{aligned}
	\end{equation}
	where $\phi$ is the parameters of our proposed ASD (Note: the parameters of ASD is fixed in this optimization objective), $w(t)$ is a scheduling coefficient, and we set it as $1-\bar{\alpha}_t$. The lower $\mathcal{L}_{\text {sp }}$ means that the stylized image generated by SPAST meets style priors from ASD, otherwise not met. \textbf{Note}: It is worth noting that our proposed style prior loss is different from SDS loss~\cite{poole2022dreamfusion} because the loss function only supports text as a reference, instead of style images. 
	To better understand the above process, consider the gradient of $\mathcal{L}_{\text {sp }}$:
	\begin{equation}
	\begin{aligned}
	\nabla_\theta \mathcal{L}_{\text {sp }}(\phi, \mathbf{x}=g(\theta))=\mathbb{E}_{t, \epsilon}[w(t) \underbrace{\left(\hat{\epsilon}_\phi\left(\mathbf{z}_{cs\_t} ; Attn(I_{s}), t\right)-\epsilon\right)}_{\text {Noise Residual }}\\ \underbrace{\frac{\partial \hat{\epsilon}_\phi\left(\mathbf{z}_{cs\_t} ; I_{s}, t\right)}{\mathbf{z}_{cs\_t}}}_{\text {U-Net Jacobian }} \underbrace{\frac{\partial \mathbf{x}}{\partial \theta}}_{\text {SPAST Jacobian }}],
	\end{aligned}
	\end{equation}
	where $\theta$ is the trainable parameters of SPAST, in such a way, the gradient of ASD is passed to SPAST by the chain rule. 
	It is noted that the timestep $t$ plays an important role in affecting the quality, and we set timestep $t=500$. 
	
	\begin{figure*}[htb]
		\includegraphics[width=2.09\columnwidth]{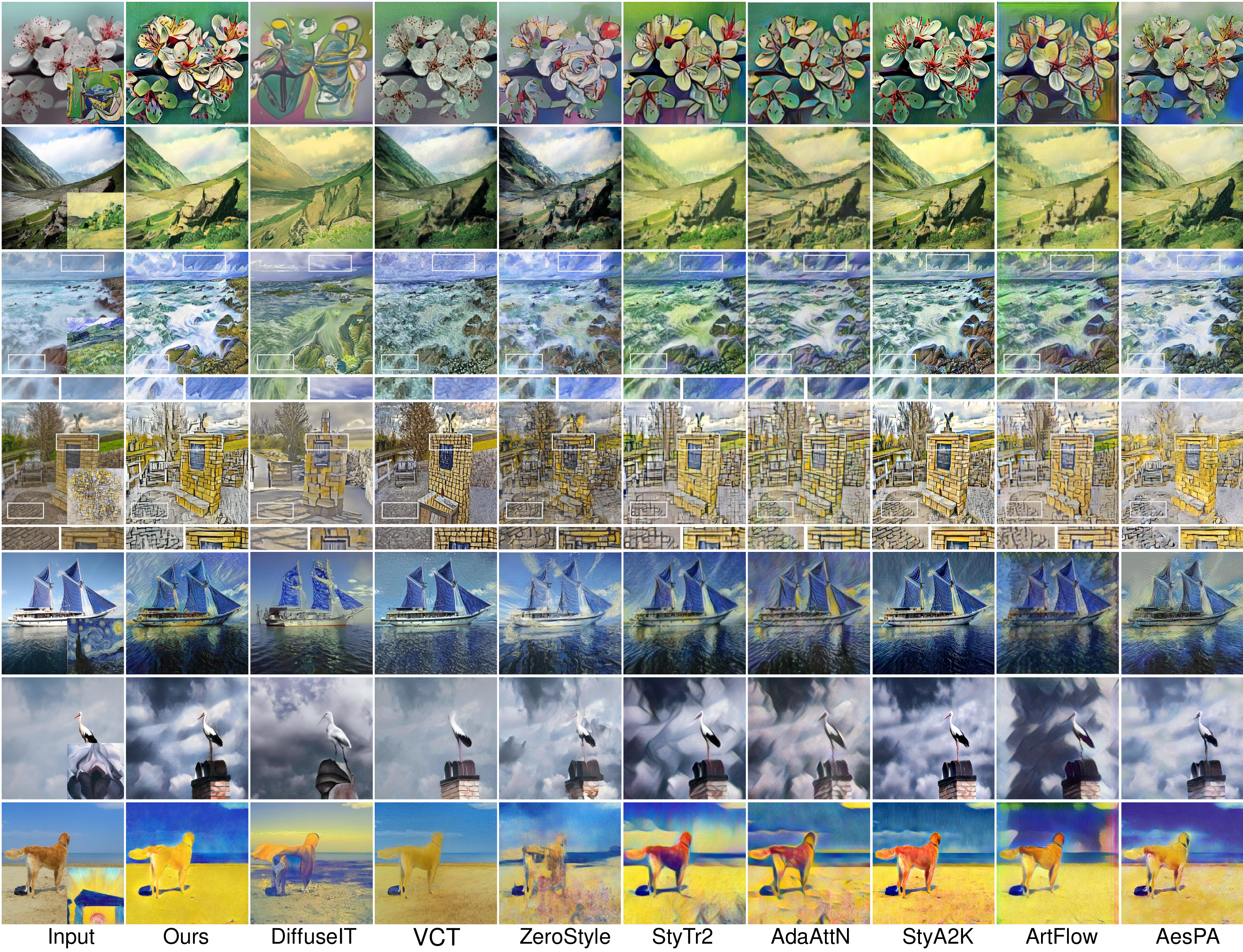} 
		\caption{Qualitative comparisons. The $1^{st}$ col shows the input images. The other columns show the stylized images generated
by other SOTA methods. Please zoom in for better observation.}
		\label{image5}
	\end{figure*}
	\begin{table*}[htb]
		\footnotesize
		\centering
		\begin{tabular}{c|c|ccccccccc}
			\hline  Method  & SPAST (Ours) & DiffuseIT & VCT & ZeroStyle & StyTr2 & AdaAttN & StyA2K& ArtFlow& AesPA \\
			\hline  Content Loss $\downarrow$ & $\mathbf{4.521}$ &8.734 & 4.846 &5.316  & 5.467 & 5.245 & 4.605&5.983&4.825 \\
			Style Loss $\downarrow$ & $\mathbf{1.0928}$ &5.1327&3.0291 & 3.2384 & 1.1496&2.8105 & 1.9832&2.1683&1.1039 \\
			Gram Loss  $\downarrow$&$\mathbf{0.00075}$&0.00162&0.00273&0.00256&0.00079&0.00135&0.000159&0.00098&0.00232\\
			LPIPS $\downarrow$ & $\mathbf{0. 401}$ &0.759 & 0.425 & 0.457 & 0.562 & 0.601 & 0.572&0.690&0.553 \\
			\hline  Deception Rate$\uparrow$ & $\mathbf{0.65}$  & 0.62 & 0.52 & 0.57 & 0.54 & 0.55 & 0.56&0.53&0.54 \\
			Preference Score & - & 45.2 & 24.5 &38.5 & 40.1 & 39.2 &  42.3 &  39.3 &  40.5\\
			\hline Inference Time $\downarrow$ & \textbf{0.010s} & 32.352s & 10m30s & 3m5s & 0.132s & 0.036s & 0.012s & 0.225s&0.038s\\
             Training Time $\downarrow$ & 57h+23h & - & - & - & 18.9h & 11.8h & 6.5h & 25.6&12.1h\\
			\hline
            
		\end{tabular}
		\caption{Quantitative comparison. The best result is signed in \textbf{bold}.}
		\label{comparisontable}
	\end{table*}
	
	\subsection{Other Loss Functions}
	\textbf{Content Loss}. Following previous style transfer methods~\cite{liu2021adaattn,deng2020arbitrary,chen2021artistic,park2019arbitrary}, we use the content loss $\mathcal{L}_{c}$ to constrain euclidean distance
	between content feature and stylized image feature which can be calculated as below:
	\begin{equation}
	\begin{aligned}
	\mathcal{L}_{cont}= \sum^{L}_{i=4}||E^{x}_{VGG}(I_{cs})
	-E^{x}_{VGG}(I_c)||_2 ,
	\end{aligned}
	\end{equation}
	where we use the $ReLU4\_1$ and  $ReLU5\_1$ to compute $\mathcal{L}_{cont}$. We use VGG to ensure that the content features of the stylized images are consistent with those of the content images. This typically means that we only require the semantic information from the content images, while the color and texture are derived from the style images. Since the shallow features of VGG (such as $ReLU1\_1$, $ReLU2\_1$, and $ReLU3\_1$) contain color and texture, we only utilize $ReLU4\_1$ and $ReLU5\_1$ to extract the content features.
	
	\textbf{Style Loss}.  Further, we use style loss $\mathcal{L}_{s}$ to constrain global style distribution between stylized images and style images: 
	\begin{equation}
	\begin{aligned}
	\mathcal{L}_{sty}= \sum^{L}_{i=1}||\mu (E^{x}_{VGG}(I_{cs})
	-\mu(E^{x}_{VGG}(I_s))||_2 \\
	+||\sigma(E^{x}_{VGG}(I_{cs}))
	-\sigma(E^{x}_{VGG}(I_s))||_2,
	\end{aligned}
	\end{equation}
	where $\mu$ and $\sigma$ denotes the channel-wise mean and standard deviation. In the experiment, we use $ReLU1\_1$, $ReLU2\_1$, $ReLU3\_1$, $ReLU4\_1$ and $ReLU5\_1$ to compute style loss $\mathcal{L}_{s}$.
	
	\textbf{Adversarial Loss}. The Generative Adversarial Network (GAN)~\cite{gulrajani2017improved,radford2015unsupervised,zhu2017unpaired,chen2021artistic,zhang2023caster}, which can effectively learn the
style distribution from style image $I_{s}$ and and push the stylized
image $I_{cs}$ look more realistic. Then, we define adversarial loss as below: 
	\begin{equation}
	\mathcal{L}_{a d v}=\underset{y \sim I_s}{\mathbb{E}}[\log (D_s(y))]+\underset{x \sim I_{c s}}{\mathbb{E}}[\log (1-D_s(x)],
	\end{equation}
	where $D_{s}$ denotes the discriminator.

	\textbf{Identity Loss}. We utilize identity loss~\cite{park2019arbitrary,lin2020tuigan,zhao2020unpaired} to preserve content structure. Following previous identity loss, $\mathcal{L}_{identity}$ can be calculated as below:
	\begin{equation}
	\begin{aligned}
	\mathcal{L}_{identity}=\lambda_{identity_{1}}(||I_{cc}-I_c||_2 +||I_{ss}-I_{s}||_2) \\
	+ \sum_{i=0}^L\lambda_{identity_{2}}(||E^{x}_{VGG}(I_{cc})-E^{x}_{VGG}(I_c)||_2\\
	+||E^{x}_{VGG}(I_{ss})-E^{x}_{VGG}(I_{s})||_2) ,
	\end{aligned}
	\end{equation}
	where $I_{cc}/I_{ss}$ denotes the image generated by SPAST with two same content/style images. We set $\lambda_{identity_{1}}=50$, $\lambda_{identity_{2}}=1$. 
	
	\subsection{Objective Loss Function}
	We summarize all the above losses to obtain the final objective loss function $\mathcal{L}$ as below:
	\begin{equation}
	\mathcal{L}=\lambda_{1}\mathcal{L}_{sty}+\lambda_{2}\mathcal{L}_{cont}+\lambda_{3}\mathcal{L}_{identity}+\lambda_{4}\mathcal{L}_{adv} + \lambda_{5}\mathcal{L}_{sp},
	\end{equation}
	where $\lambda_{1}$, $\lambda_{2}$, $\lambda_{3}$, $\lambda_{4}$ and $\lambda_{5}$ are hyper-parameters to adjust balance of each loss term.

	
	\subsection{Why does the LGWSSM work?}
	
\begin{figure}[htbp]
		\centering
		\includegraphics[width=1\columnwidth]{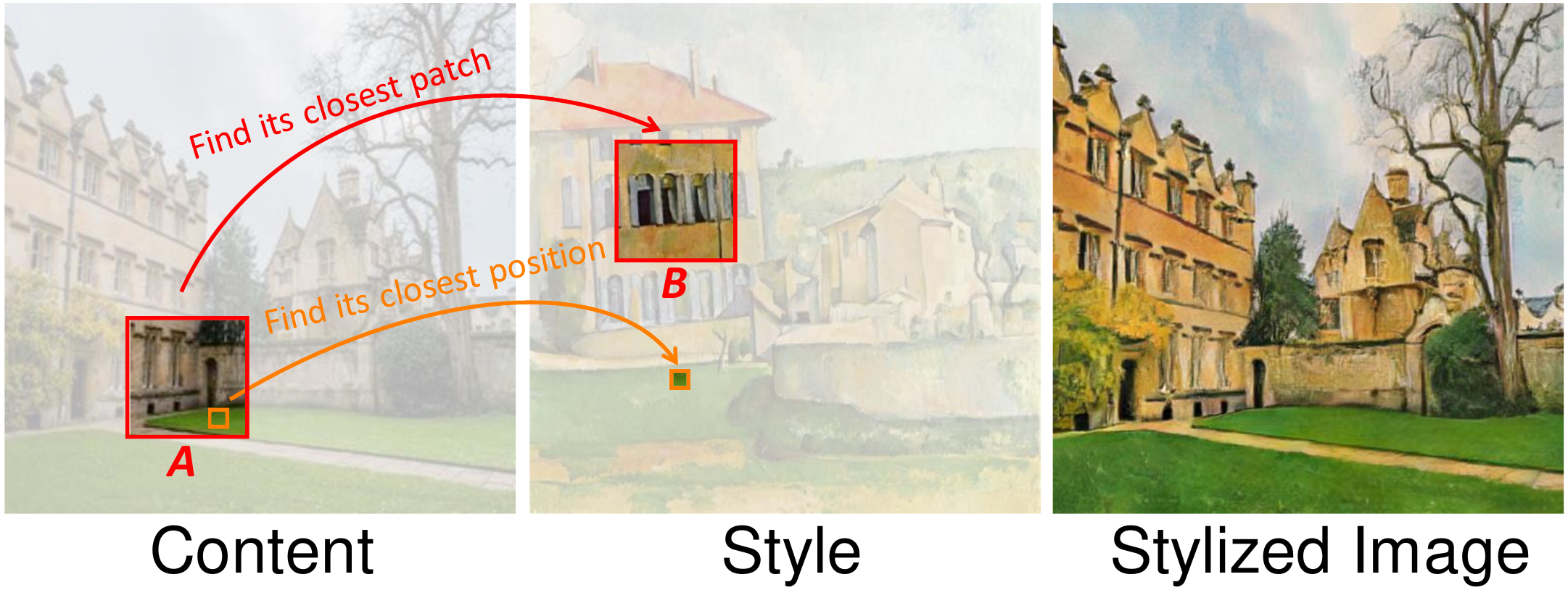} 
		\caption{Illustration about why the Local-global Window Size Stylization Module works.}
		\label{image9}
	\end{figure}

	Some existing methods~\cite{park2019arbitrary,li2023rethinking,zhu2023all} utilize attention to either
consider the position of content features or the region of content
features to match the position and region of their closest
style features. Although such a way effectively enables style
transfer, they fail to consider content-style relationships from
both position and region perspective. To this end, we first
propose LWSSM to make each region of the content feature
find its closest region of the style feature (i.e., \protect\textcolor{red}{Red} box in Fig.~\ref{image9}) and learn region-wise style information. Secondly, we argue that although LWSSM is effective in learning style information from the region perspective, it doesn't work in some cases. As shown in Fig.~\ref{image9}, given content region $A$ and its closest style region $B$, some positions in $A$ cannot find their closest position from $B$. To this end, we propose GWSSM to push each position of the content feature region to find its closet position from the outside of the region $B$ (i.e., \protect\textcolor{orange}{Orange} box in Fig.~\ref{image9}). 
	
	\subsection{The analysis of Timestep $t$ in $\mathcal{L}_{sp}$}
\begin{figure}[htb]
		\centering
		\includegraphics[width=1\columnwidth]{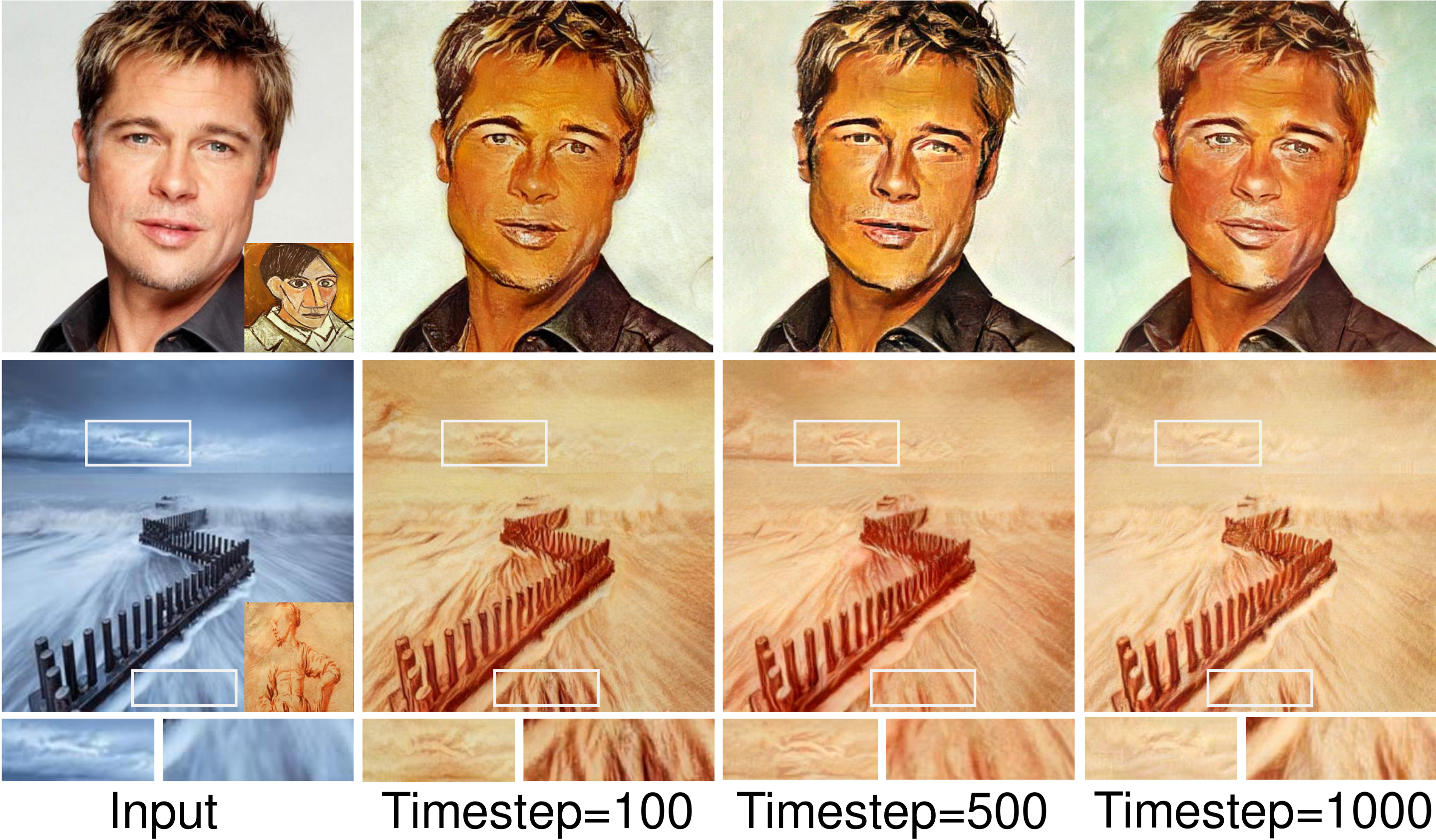} 
		\caption{Stylized images by our proposed SPAST. It is worth noting that the timestep $t$ of $\mathcal{L}_{sp}$ is the key to affect the stylized image. We show stylized samples with different timestep $t$ (e.g., Timestep=100/500/1000).}
		\label{image6}
	\end{figure}

	Previous methods~\cite{zhang2023prospect,agarwal2023image,zhang2024towards} have verified that diffusion models can be roughly divided into three stages based on noise levels: high, medium, and low. The high noise stage primarily affects the content structure of the generated images, the medium noise stage influences both the content structure and style patterns of the generated images, while the low noise stage primarily impacts the style patterns. Inspired by this, we propose the $\mathcal{L}_{sp}$ loss function. It is worth noting that we find that timestep $t$ of $\mathcal{L}_{sp}$ is an important role in affecting the quality of stylized images and show stylized samples with different timestep in Fig.~\ref{image6}. More timesteps may smooth the stylized image generated by SPAST and remove the local details of stylized images. 
    
	\section{Experiments}
	\subsection{Implementation Details}
	\textbf{Stage One}
	In this stage, we select 79,433 style images from WikiArt~\cite{wikiart} as the training set. For each training step, we randomly select style image samples and resize them to $512\times512$ pixels. We train our proposed ASD for 300,000 iterations on Two NVIDIA RTX 3090 GPUs via Pytorch 1.11~\cite{paszke2019pytorch} and Adam optimizer~\cite{kingma2014adam} with a learning rate of 0.000001.
	
	\textbf{Stage Two}
	We select the content images and
style images from MS-COCO~\cite{lin2014microsoft} and WikiArt~\cite{wikiart} as the training set. During training, all 82,783 content images and
79,433 style images are resized to $512\times512$ pixels and they are cropped to $256\times256$ pixels. To feed stylized image into ASD to calculate Eq.~\ref{sp}, the stylized images are resized to $512\times512$ pixels. We set $\lambda_{1}=1$, $\lambda_{2}=1$, $\lambda_{3}=1$, $\lambda_{4}=1$, and $\lambda_{5}=1$. 
	With a learning rate of 0.0001 and a batch of 1, we train our proposed SPAST for 160,000 iterations upon the Pytorch 1.11~\cite{paszke2019pytorch} and Adam optimizer~\cite{kingma2014adam} with a learning rate of 0.0001 on a single NVIDIA RTX 3090 GPU.
	
	\subsection{Comparison with SOTA methods.}
	As shown in Fig.~\ref{image5}, we compare the proposed method with the SOTA methods, including ZeroStyle~\cite{deng2024z}, StyTr2~\cite{deng2022stytr2}, AdaAttN~\cite{liu2021adaattn}, StyA2K~\cite{zhu2023all}, ArtFlow~\cite{an2021artflow}, AesPA~\cite{hong2023aespa}, DiffuseIT~\cite{kwon2023diffusion} and VCT~\cite{cheng2023general}. 
	
	\subsection{Qualitative Results.}
	We show qualitative stylized results from different style transfer methods, including SSM and pre-trained large-scale LMM. The images used for the inference are collected from the test cases provided by some existing open-source style transfer methods~\cite{park2019arbitrary,wang2022aesust}.
    As representatives of the former, the color of stylized image from StyTr2~\cite{deng2022stytr2} are deviated from style image (e.g., $1^{st}$, $3^{rd}$ and $7^{th}$ rows). AdaAttN~\cite{liu2021adaattn} suffers from the content loss in detail (e.g., $4^{th}$ and $5^{th}$ rows). StyA2K~\cite{zhu2023all} shows less stylized with limited colors and textures (e.g., $4^{th}$ and $7^{th}$ rows). ArtFlow~\cite{an2021artflow} often brings unwanted local texture (e.g., $3^{rd}$, and $5^{th}$ rows). AesPA~\cite{hong2023aespa} fails to keep color consistency between stylized image and style image (e.g., $1^{st}$, and $5^{th}$ rows). As representatives of the latter, ZeroStyle~\cite{deng2024z} tends to introduce obvious artifacts and abrupt color (e.g., $1^{st}$ and $7^{th}$ rows). DiffuseIT~\cite{kwon2023diffusion} fails to preserve the content structure (e.g., $1^{st}$ and $6^{th}$ rows). VCT~\cite{cheng2023general} has limitations in learning local patterns and global distribution from style images (e.g., $1^{st}$ and $6^{th}$ rows).

	\textbf{Quantitative Comparisons.}
	In line with previous style
transfer methods such as those in references~\cite{zhu2023all,deng2022stytr2}, the
average content loss is employed to gauge the preservation
of content structure. The average style loss and gram loss are
utilized to measure the global style distribution. A smaller
value indicates better preservation of content structure or
style distribution. Additionally, we calculate the LPIPS metric~\cite{zhang2018unreasonable} to evaluate the content consistency between stylized
images and input content images. Again, a smaller value
implies better content consistency of the stylized images.
In this section, 20 content images and 30 style images are
used to generate 600 stylized images for both the state-of-the-
art (SOTA) methods and our proposed approach. As
demonstrated in Tab. 1, our proposed SPAST surpasses the
existing SOTA methods.
	\begin{figure*}[htb]
		\centering
		\includegraphics[width=2.1\columnwidth]{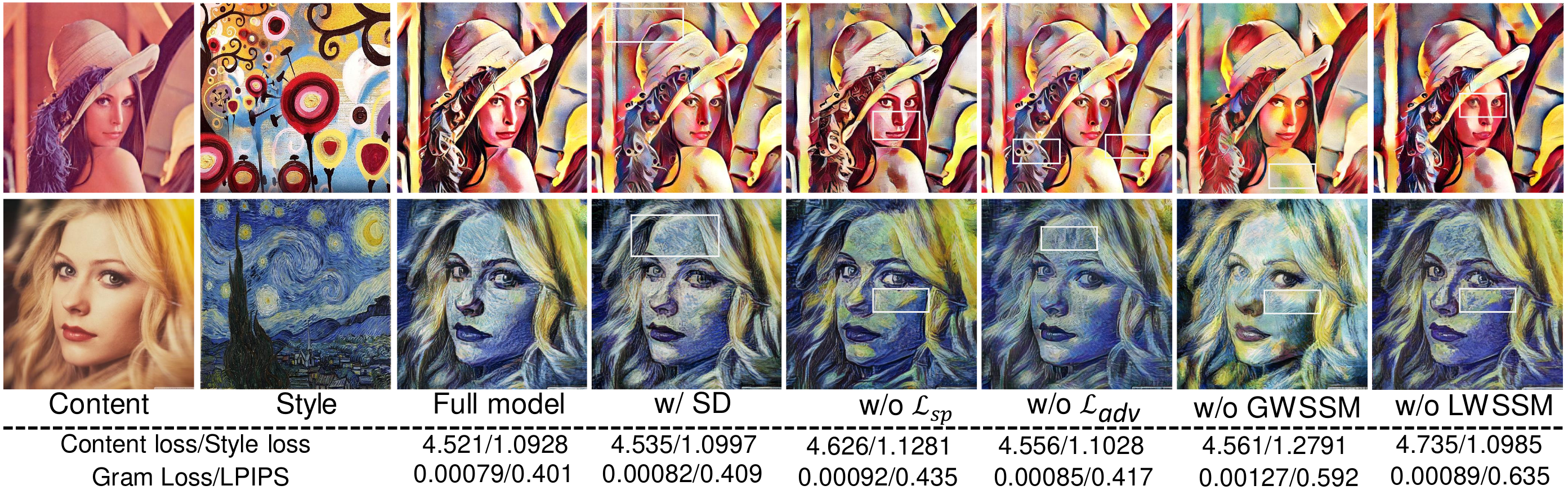} 
		\caption{Ablation study.}
		\label{ablation}
	\end{figure*}
    \begin{figure}[htb]
		\centering
		\includegraphics[width=1\columnwidth]{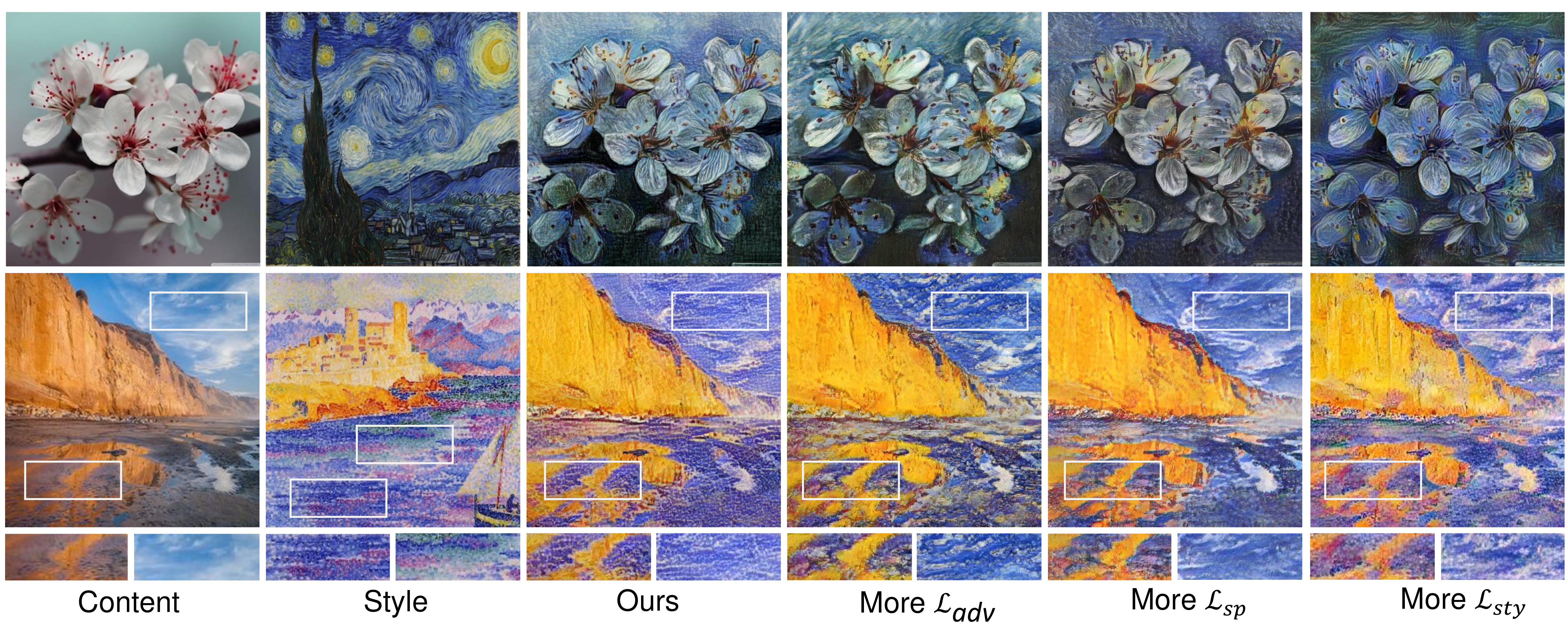} 
		\caption{An ablation study on how the choice of loss function affects the  stylized images.  Please zoom-in for better comparison.}
		\label{image14}
	\end{figure}
	
	\textbf{User Study.}
	Deception scores indicate a user’s ability to
distinguish between stylized and human-created art images.
A higher score implies a greater percentage of stylized
images being misidentified as human-created art images. For
each method we randomly chose 20 synthesized images and
had 50 subjects make guesses. Since certain methods may excel at generating specific stylized images, to enhance fairness, the 20 synthesized images generated by each method are derived from different content images and style images. We also randomly selected
an equal number of WikiArt images and asked the same 50
subjects to assess them.
	
	Preference Score~\cite{wang2022aesust,chen2021dualast} conducts A/B user studies to assess the
stylized effects of our proposed SPAST in comparison with
the SOTA methods. A total of 100 content-style pairs are
selected to generate 10,000 stylized images for each method.
Subsequently, in each pair, aside from the content and style images, two stylized images are displayed – one generated by
SPAST and the other by a randomly chosen SOTA method.
The participants are then asked to select their preferred
stylized image.We ultimately collected 5,000 votes from 50
participants and presented the percentage of votes for each
method in the sixth row of Tab.~\ref{comparisontable}. 
As indicated in the $7^{th}$ row and $3^{rd}$ col in Tab.~\ref{comparisontable}, 45.2 means that 45.2\% of participants favor the images generated by DiffuseIT, while 54.8\% of participants prefer the images produced by our
proposed method.
	
	\textbf{Efficiency Comparison.}
	We evaluate inference time~\cite{zhang2025lgast,zhang2024rethink} of our proposed SPAST and other SOTA methods in Tab.~\ref{comparisontable}. We implement SPAST on a single NVIDIA RTX 3090 GPU for inference under 512$\times$ 512 resolution. 

    We also evaluate the training time of our proposed SPAST and other SOTA methods in Tab.~\ref{comparisontable}. We retrain SPAST and other SOTA methods on a single NVIDIA RTX 3090 GPU under 512$\times$ 512 resolution with a batch size 1. In stage one, we train SPAST for 300,000 iterations (costing 57 hours). Similarly, SPAST is also trained for 160,000 iterations (costing 23 hours) in stage two. We also train other state-of-the-art (SOTA) methods for  160,000 iterations. For ZeroStyle~\cite{deng2024z}, DiffuseIT~\cite{kwon2023diffusion}, and VCT~\cite{cheng2023general}, these methods require online optimization when applying style information from a style image to a content image. As a result, they do not support arbitrary style transfer, so we do not provide their training times.

	\subsection{Ablation Studies.}
	\textbf{Analysis of proposed module.} As introduced in Sec.~\ref{lgwssm}. The LGWSSM comprises of the LWSSM and GWSSM. To verify their effect, we retrain our proposed SPAST with and without LWSSM/GWSSM in Fig.~\ref{ablation}. Without LWSSM, the model neglects the local style patterns of the style image. Without GWSSM, the color of the stylized image deviates from the style image. The above analysis is supported by reported quantitative results in Fig.~\ref{ablation}.
	
	\textbf{Analysis of proposed loss.} To investigate the influence of the proposed style prior loss $\mathcal{L}_{sp}$ and adversarial loss $\mathcal{L}_{adv}$, we remove them from our model and show the experimental results in Fig.~\ref{ablation}. Without $\mathcal{L}_{sp}$, the stylized image show disharmonious patterns and pattern artifacts. ``w/ SD" means to retrain our proposed SPAST with Stable Diffusion as style priors, the stylized image introduces disharmonious patterns. Thus, our proposed $\mathcal{L}_{sp}$ can help effectively improve the stylization quality of the small model-based method. Without $\mathcal{L}_{adv}$, the stylized images show abrupt colors. As shown in Fig.~\ref{image14}, we adjust the weights of the $\mathcal{L}_{adv}$, $\mathcal{L}_{sp}$, and $\mathcal{L}_{sty}$ functions to verify the contributions of these three loss functions and to illustrate how we select their weights. We  don’t adjust the weights of $\mathcal{L}_{cont}$ and $\mathcal{L}_{identity}$ here because these two loss functions are used to preserve the content structure of the generated images and their weights, are set according to the configuration used in SaNet~\protect\cite{park2019arbitrary}. First, we increase the weight of the $\mathcal{L}_{adv}$ by setting $\lambda_{4}=5$. As shown in Fig.~\ref{image14}, we can see that the stylized images show the mussy texture (see $1^{st}$ row and $4^{th}$ col). Next, we increased the weight of the $\mathcal{L}_{sp}$ loss function by setting $\lambda_{5}=5$, and we find that with more $\mathcal{L}_{sp}$, the colors of the stylized image tends to deviate from the style image (see $1^{st}$ row and $5^{th}$ col). Finally, increasing the weight of the $\mathcal{L}_{sty}$ results in the stylized image failing to capture the detailed textures of the style image (see $2^{nd}$ row and $6^{th}$ col).
	\section{Conclusion}
	We proposed a novel method, called SPAST, which can
render a high-quality stylized images with better content
structure preservation and less inference time. The extensive experiments demonstrate that our proposed SPAST outperforms
SOTA methods.
	The extensive experiments demonstrate that our proposed SPAST outperforms state-of-the-art style transfer methods. 

\section{CRediT authorship contribution statement}
\textbf{Zhanjie Zhang:} Conceptualization, Methodology, Software, Writing – original draft.
\textbf{Quanwei Zhang:} Conceptualization, Methodology, Writing – original draft.
\textbf{Junsheng Luan:} Conceptualization, Methodology,
Writing – review editing. \textbf{Mengyuan Yang}: Software, Investigation, Data
curation, Validation, Writing – review editing. \textbf{Yun Wang:}
Software, Validation, Data curation, Writing – review editing.
\textbf{Lei Zhao:} Supervision, Writing – review editing.

\section{Declaration of Competing Interest}
The authors declare that they have no known competing financial interests or personal relationships that could have appeared to influence the work reported in this paper.

\section{Data availability}
Data will be made available on request.

\section{Acknowledgments}
This work was supported in part by Zhejiang Province Program (2024C03263, LZ25F020006), the National Program of China (62172365, 2021YFF0900604, 19ZDA197), Macau project: Key technology research and display system development for new personalized controllable dressing dynamic display, Ningbo Science and Technology Plan Project (2022Z167, 2023Z137), and MOE Frontier Science Center for Brain Science \& Brain-Machine Integration (Zhejiang University).

\bibliographystyle{cas-model2-names}

\bibliography{cas-refs}


\end{document}